%% file: main.tex
\documentclass[conference, onecolumn]{IEEEtran}
\usepackage[english]{babel}
\usepackage{comment}
\usepackage{graphicx}  
\usepackage{amssymb,amsfonts}
\usepackage[fleqn]{amsmath}
\usepackage{subfig}
\usepackage{url}

\newcommand{\method}{$M^3Fusion$}

\begin{document}
\title{\method: A Deep Learning Architecture for Multi-\{Scale/Modal/Temporal\} satellite data fusion}

\author{P. Benedetti,
		D. Ienco,
        R. Gaetano,
        K. Os\'{e},
        R. Pensa 
        and S. Dupuy}
        
\thanks{ P. Benedetti , K. Os\'{e} are with UMR-TETIS laboratory, IRSTEA, University of Montpellier, Montpellier, France (email: paola.benedetti@irstea.fr; kenji.ose@irstea.fr)}
\thanks{D. Ienco is with UMR-TETIS laboratory, IRSTEA, University of Montpellier, Montpellier, France and with LIRMM laboratory, Montpellier, France (email: dino.ienco@irstea.fr).}%
\thanks{R. Gaetano is with CIRAD, UMR TETIS, 500 Rue J.-F. Breton, F-34000 Montpellier, France and with UMR TETIS, Univ. Montpellier, AgroParisTech, CIRAD, CNRS, IRSTEA, Montpellier, France (email: raffaele.gaetano@cirad.fr)}
\thanks{S. Dupy is with CIRAD, UMR TETIS, F-97410 Saint-Pierre, Réunion, France and UMR TETIS, Univ. Montpellier, AgroParisTech, CIRAD, CNRS, IRSTEA, Montpellier, France (email: stephane.dupuy@cirad.fr)}

\maketitle

\begin{abstract}
Modern Earth Observation systems provide sensing data at different temporal and spatial resolutions.  Among optical sensors, today the Sentinel-2 program supplies high-resolution temporal (every 5 days) and high spatial resolution (10m) images that can be useful to monitor land cover dynamics. On the other hand, Very High Spatial Resolution images (VHSR) are still an essential tool to figure out land cover mapping characterized by fine spatial patterns. Understand how to efficiently leverage these complementary sources of information together to deal with land cover mapping is still challenging.

With the aim to tackle land cover mapping through the fusion of multi-temporal High Spatial Resolution and Very High Spatial Resolution satellite images, we propose an End-to-End Deep Learning framework, named \method{}, able to leverage simultaneously the temporal knowledge contained in time series data as well as the fine spatial information available in VHSR information.
Experiments carried out on the \textit{Reunion Island} study area asses the quality of our proposal considering both quantitative and qualitative aspects. 
\end{abstract}


\section{Introduction}
\label{sec:intro}
\input{intro}

\section{Data}
\label{sec:data}
\input{data}

\section{Contributions}
\label{sec:method}
\input{method}

\section{Experiments}
\label{sec:expe}
\input{experiments}

\section{Conclusions}
\label{sec:conclu}
In this article, we proposed a new deep learning architecture for the fusion of satellite data with high temporal/spatial resolution and very high spatial resolution to perform land use mapping. Experiments carried out on a real study site  validate the quality and the results of our approach compared to the ones obtained by a common machine learning methods usually employed in the field of remote sensing for the same task. In the future, we plan to study the extension of our architecture to take into account other complementary data sources.

\section{Acknowledgements}
This work was supported by the French National Research Agency under the Investments for the Future Program, referred as ANR-16-CONV-0004 and the GEOSUD project with reference ANR-10-EQPX-20, as well as from the financial contribution from the Ministry of Agriculture's "Agricultural and Rural Development" trust account. This work also used an image acquired under the CNES Kalideos scheme (La R\'{e}union site).

\bibliography{m3fusion}
\bibliographystyle{IEEEtran}
\end{document}

%% file: intro.tex
Modern Earth Observation systems produce huge volumes of data every day. This information can be organized into time series of high-resolution satellite imagery (SITS) (i. e. Sentinel) that are useful for area monitoring over time.

In addition to this high temporal frequency information, we can also obtain Very High Spatial Resolution (VHSR) information, such as Spot6/7 or Pleiades imaging, with a more limited temporal frequency~\cite{MaggioriTCA17} (e. g. once a year).

The analysis of time series and its coupling/fusion with punctual VHSR data remains an important challenge in the field of remote sensing.~\cite{Karpatne16,Schmitt17}.

In the context of land use classification, employing high spatial resolution (HSR) time series, instead of a single image of the same resolution, can be useful to distinguish classes according to their temporal profiles~\cite{Abade15}. On the other hand, the use of fine spatial information helps to differentiate other kind of classes that need spatial context information at higher scale~\cite{Schmitt17}. 

Typically, the approaches that use these two types of information~\cite{IngladaVATMR17,LebourgeoisDVAB17}, perform data fusion at descriptor level~\cite{Schmitt17}. This type of fusion involves extracting a set of independent features for each data source (time series, VHSR image) and then stacking these features together to feed a traditional supervised learning method (i. e., Random Forest).

Recently, the deep learning revolution~\cite{Zhang16} has shown that neural network models are well adapted tools for automatically managing and classifying remote sensing data~\cite{Zhang16}. 
The main characteristic of this type of model is the ability to simultaneously extract features optimized to image classification and the associated classifier.
This advantage is fundamental in a data fusion process such as the one involving high resolution time series (i. e. Sentinel-2) and VHSR data (i. e. Spot6/7 and/or Pleiades).

Considering deep learning methods, we can find two main families of approaches: convolutional neural networks~\cite{Zhang16} (CNN) and recurrent neural networks~\cite{BengioCV13} (RNN).  
CNN are well suited to model the spatial autocorrelation available in an image, while RNN networks are especially tailored to manage time dependencies ~\cite{IencoGDM17} from multidimensional time series.

In this article, we propose to leverage both CNN and RNN to address the fusion problem between an HSR time series of Sentinel-2 images and a VHSR image on the same study area with the goal to perform land use mapping.
The method we propose, named \method{} (Multi-Scale/Modal/Temporal Fusion), consists in a deep learning architecture that integrates both a CNN component (to manage VHSR information) and an RNN component (to analyze HSR time series information) in an end-to-end learning process. 
Each information source is integrated through its dedicated module and the extracted descriptors are then concatenated to perform the final classification. 
Setting up such a process, which takes both data sources into account at the same time, ensures that we can extract complementary and useful features for land use mapping.

To validate our approach, we conducted experiments on a data set involving the Reunion Island study site. This site is a French Overseas Department located in the Indian Ocean (east of Madagascar) and it will be described in Section ~\ref{sec:data}.
The rest of the article is organized as follows: Section~\ref{sec:method} introduces the \method Deep Learning Architecture for the multi-source classification process. The experimental setting and the findings are discussed in Section~\ref{sec:expe} and conclusions are drawn in Section~\ref{sec:conclu}.

%% file: data.tex
The study was carried out on Reunion Island, a French overseas department located in the Indian Ocean. The dataset consists of a time series of 34 Sentinel-2 (S2) images acquired between April 2016 and May 2017, as well as a very high spatial resolution image (VHSR) SPOT6/7 acquired in April 2016 and covering the whole island. The S2 images used are those provided at level 2A by the Continental Surfaces pole THEIA\footnote{Données disponibles via \url{http://theia.cnes.fr}, pre-treated in surface reflectance via the \textit{MACCS-ATCOR Joint Algorithm}~\cite{Hagolle2015} developed by the National Centre for Space Studies (CNES).}, where the bands at 20~m resolution were resampled to 10~m. A preprocessing was performed to fill cloudy observations through a linear multi-temporal interpolation over each band (cfr. \textit{Temporal Gapfilling}, \cite{IngladaVATMR17}), and six radiometric indices were calculated for each date (NDVI, NDWI, brightness index - BI, NDVI and NDWI of infrared means - MNDVI and MNDWI, and vegetation index Red-Edge - RNDVI) \cite{IngladaVATMR17,LebourgeoisDVAB17}). A total of 16 variables (10 surface reflectances plus 6 indices) are considered for each pixel of each image in the time series.

The SPOT6/7 image, originally consisting of a 1.5~m panchromatic band and 4 multispectral bands (blue, green, red and near infrared) at 6~m resolution, was merged to produce a single multispectral image at 1.5~m resolution and then resampled at 2~m because of the network  architecture learning requirements.\footnote{This was done to ensure a direct and non-overlapping correspondence between the time series pixels (10~m) and a block of VHSR pixels (5 $\times$ 5).}. Its final size is 33280 $\times$ 29565 pixels on 5 bands (4 reflectors \textit{Top of Atmosphere} plus the NDVI). This image was also used as a reference to realign the different images in the time series by a searching and mapping anchor points, in order to improve the spatial coherence between the different sources.

The field database was built from various sources: (i) the graphical parcel register (RPG) data set of 2014, (ii) GPS records from June 2017 and (iii) photo interpretation of the VHSR image conducted by an expert, with knowledge of the territory, for natural and urban spaces . All polygon contours have been resumed using the VHSR image as a reference. The final dataset includes a total of 322\,748 pixels (2\,656 objects) distributed over 13 classes, as indicated in the Table \ref{tab:data}.

\begin{table}[!ht]
\centering
\begin{tabular}{|l||c|c|c|}
	\hline
\textbf{Class} & Label & \# \textbf{Objects} & \# \textbf{Pixels} \\ 
\hline \hline
0 & {\em Crop Cultivations} & 380 & 12090 \\ \hline
1 & {\em Sugar cane} & 496 & 84136 \\ \hline
2 & {\em Orchards} & 299 & 15477 \\ \hline
3 & {\em Forest plantations} & 67 & 9783 \\ \hline
4 & {\em Meadow} & 257 & 50596 \\ \hline
5 & {\em Forest} & 292 & 55108 \\ \hline
6 & {\em Shrubby savannah} & 371 & 20287 \\ \hline
7 & {\em Herbaceous savannah} & 78 & 5978 \\ \hline
8 & {\em Bare rocks} & 107 & 18659 \\ \hline
9 & {\em Urbanized areas} & 125 & 36178 \\ \hline
10 & {\em Greenhouse crops}& 50 & 1877 \\ \hline
11 & {\em Water Surfaces} & 96 & 7349 \\ \hline
12 & {\em Shadows} & 38 & 5230 \\ \hline
\end{tabular}
\caption{Characteristics of the Reunion Dataset\label{tab:data}}
\end{table}

%% file: method.tex
\begin{figure*}[t]
\centering
\includegraphics[scale=0.5]{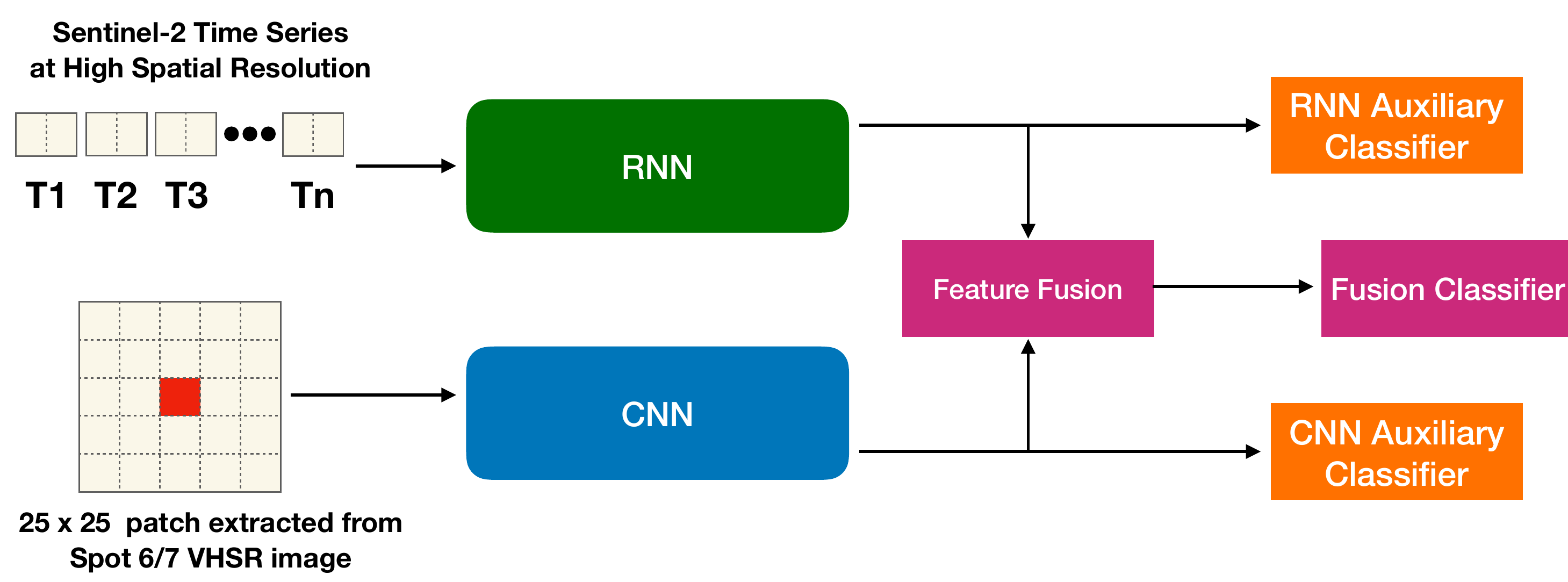}
\caption{ \label{fig:overview}Visual representation of \method{}.}
\end{figure*}

\subsection{\method{} model description }
Figure~\ref{fig:overview} visually describes the  \method{} approach proposed in this article. First of all, we define the input data for our deep learning model. \method{} takes as input a data $\{ (x_i, y_i)\}_{i=1}^{M}$ where each example is associated with a class value $y_i \in {1,...,C}$. An example $x_i$ is defined as a pair $x_i = (ts_{i}, patch_{i})$ such that $ts_{i}$ is the (multidimensional) time series of a  Sentinel-2 pixel (10m resolution) and $patch_{i}$ is a subset of the image Spot6/7 (2m resolution) centered around the corresponding pixel Sentinel-2. Every example has two representation: a temporal HSR modality (provided by the Sentinel-2 time series) and a VHSR modality (provided by the Spot6/7 image).

For every $patch_{i}$, we fixed the window size to 25$\times$25 pixels on the Spot6/7 (which correspond to a window size 5x5 on a Sentinel-2 image) centered around a Sentinel-2 pixel described by the corresponding $ts_{i}$. 

In order to merge the temporal information with the VHSR one contained in the Spot 6/7 image, we designed a network which has two parallel branches, one for each of the two modes (spatial/temporal). For the time series concerning the Sentinel-2 pixel we use a Recurrent Neural Network (RNN) architecture. In particular, we used a Gated Recurrent Unit (GRU) introduced in~\cite{ChoMGBBSB14} which has already demonstrated its effectiveness in the remote sensing field~\cite{MouGZ17,Dinh18}. On the other hand, the spatial information, with a scale of 2m, introduced through VHSR image is integrated through the use of a Convolutional Neural Network~\cite{MaggioriTCA17} which allows to extract spatial context knowledge around the Sentinel-2 pixel.

Via the two streams of analysis we learn two complementary groups of features that we successively leverage for the classification that is performed at the scale of the Sentinel-2 pixel. According to the philosophy introduced in~\cite{HouLW17}, the proposed architecture aims to learn two sets of complementary features (thanks to the different spatial and temporal modalities) that are as much as possible discriminative when used alone. To ensure this last point, the strategy provides in~\cite{HouLW17} introduces two additional auxiliary classifiers, working independently on each group of features, as shown in the Figure~\ref{fig:overview}. A third classifier, working on the fusion (by concatenation) of the two sets of features, perform the final land use mapping.

Each of the above mentioned classifiers is realized by directly connecting the associated features to the output neurons on which SoftMax activation function is successively applied~\cite{Zhang16}. The model weights are learned by back-propagation. The cost-function associated to the model is derived by a linear combination of the individual cost function of each of the classifiers.

\subsection{Integration of information from the HSR time series}

Recently, recurrent neural network (RNN) approaches demonstrated their quality in the remote sensing field to produce land use mapping using time series of optical images~\cite{IencoGDM17} and recognize vegetation cover status using Sentinel-1 radar time series~\cite{Dinh18}. Motivated by these recent works, we decided to introduce an RNN module to integrate information from the Sentinel-2 time series into our fusion process. In our model we chose the GRU unit (Gated Recurrent Unit) introduced by~\cite{ChoMGBBSB14}, coupled with an \textit{attention} mechanism ~\cite{BritzGL17}. Attention mechanisms are widely used in automatic signal processing (language or 1D signal) and they allow to combine together the information extracted by the GRU model at the different timestamps.
The input of a GRU unit is a sequence ($x_{t_1}$,..., $x_{t_N}$) where a generic element $x_{t_i}$ is a multidimensional vector and $t_i$ refers to the corresponding date in the time series. The output returned by the GRU model is a sequence of feature vectors learned for each date: ($h_{t_1}$,..., $h_{t_N}$) where each $h_{t_i}$ has the same dimension $d$. Their matrix representation $H \in \mathbb{R}^{N,d}$ is obtained  vertically stacking the set of vectors.
The attention mechanism allows to combine together these different vectors $h_{t_{i}}$, in a single one $rnn_{feat}$, to better combine the information returned by the GRU unit at each of the different timestamps. The attention formulation we used, from a vector sequence of the learned descriptors ($h_{t_1}$,..., $h_{t_N}$), is the following one:
\begin{align}
v_{a} &= tanh( H \cdot W_{a} + b_{a}) \tag{1} \\
\lambda &= SoftMax(v_{a} \cdot u_{a}) \tag{2} \\
rnn_{feat} &= \sum_{i=1}^{N} \lambda_i \cdot h_{t_{i}} \tag{3}
\end{align}
Matrix $W_{a} \in \mathbb{R}^{d,d}$ and vectors $b_{a}, u_{a} \in \mathbb{R}^{d}$ are parameters learned during the process. These parameters allow to combine the vectors contained in matrix $H$.
The purpose of this procedure is to learn a set of weights ($\lambda_{t_1}$,..., $\lambda_{t_N}$) that allows the contribution of each timestamp to be weighted $h_{t_i}$ through a linear combination. The $SoftMax(\cdot)$~\cite{IencoGDM17} function is used to normalize weights $\lambda$ so that their sum is equal to 1.
The output of the RNN module is the feature vector $rnn_{feat}$, these features encode temporal information related to $ts_i$ for the pixel $i$.

\subsection{ Integration of VHSR information}

\begin{figure}[ht!]
\centering
\includegraphics[width=.72\columnwidth]{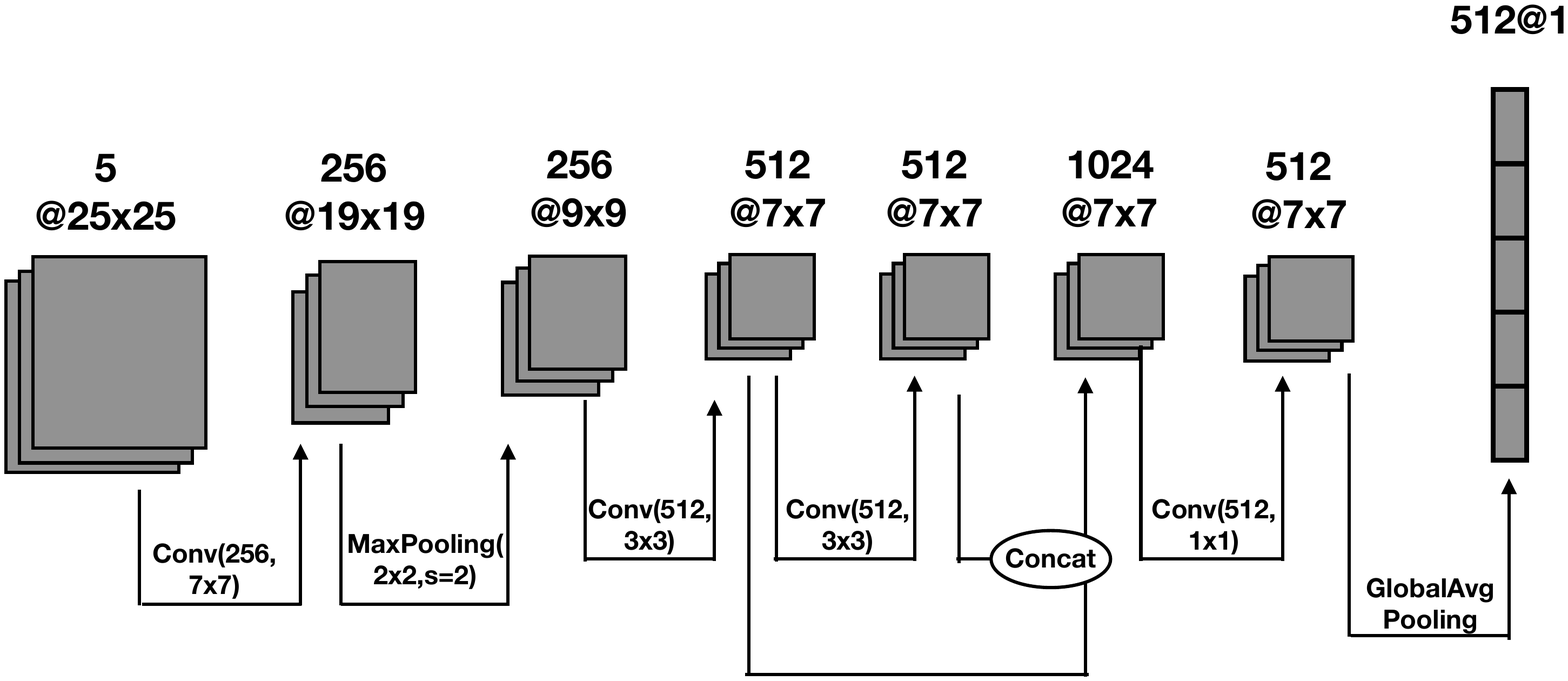}
\caption{ \label{fig:sCNN} Convolutional Neural Network Structure }
\end{figure}

The VHSR information is integrated in \method{} through a CNN module. Computer vision literature offers several convolutional architectures~\cite{HeZRS16,HuangLMW17}. Most of these networks are designed to process RGB images (three channels) having size higher then 200x200. Such networks are composed by multiple layers ( tens or hundreds ). In our scenario, the image patches to analyze have a size of 25x25 and they are described by five channels. In order to propose a CNN module that well fit our scenario and remains computational affordable parameters-wise, we design the CNN module depicted in Figure~\ref{fig:sCNN}.

\vspace{-1cm}

Our CNN network applies a first kernel 7$\times$7 to the five-channel patch to produce 256 feature maps. Then, a \textit{max pooling} layer is used to reduce the size and the number of parameters. Two successive convolution operations with a kernel 3$\times$3 extract 512 features maps each, which are then concatenated and reduced again by a convolution kernel 1$\times$1 (final size 512$\times$7$\times$7). 
Finally, a Global Average Pooling operation allows to produce a vector of features with size equals to 512. 

Each convolution is associated with a linear filter, followed by a Rectifier Linear Unit activation function~\cite{NairH10} to introduce non-linearity and a Batch Normalization step~\cite{IoffeS15}. The key points of our proposal are twofold: a) a higher number of filters in the first step and b) the concatenation of features maps at different resolutions.
The first point is related to the higher amount of spectral input information (five channels) compared to RGB images. To better exploit the high spectral richness of this data, we have increased the number of feature maps generated at this stage. The second point concerns the concatenation of features maps. With the goal to exploit information at different resolutions we adopt a philosophy similar to~\cite{HuangLMW17}. 
The output of this module is a vector of dimensions 512 ($cnn_{feat}$) which summarizes the spatial context ($patch_{i}$) associated to the Sentinel-2 pixel $i$.

\subsection{ Merging descriptors through an End-To-End process }
One of the advantages of deep learning, compared to standard machine learning methods, is the ability to link, in a single pipeline, the feature extraction step and the associated classifier~\cite{Zhang16}. This quality is particularly important in a multi-source, multi-scale and multi-temporal fusion process. \method{} leverages this asset to be able to extract complementary descriptors from two sources of information that describe the same pixel from different viewpoints. In order to further strengthen the complementarity and then, the discriminative power of the learned features for each information stream, we adapted the technique proposedin~\cite{HouLW17} to our problem. In~\cite{HouLW17}, the authors proposed to learn two complementary representations (using two convolutional networks) of the same image. The discriminative power is enhanced by two auxiliary classifiers, linked to each group of features, in addition to the classifier that uses the merged information through a sum operation.
In our case, we have two complementary sources of information (sentinel-2 time series and VHSR data) to which two auxiliary classifiers are attached in order to independently increase their ability to recognize land cover classes. Regarding the classifier that exploits the full set of features, we feed it concatenating the output features of both CNN ($cnn_{feat}$) and RNN ($rnn_{feat}$) module together.
The learning process will involve optimizing three classifiers at the same time, one specific to $rnn_{feat}$, a second one related to $cnn_{feat}$ and the third one that consider $[rnn_{feat},cnn_{feat}]$. 

The cost function associated to our model is :
\begin{align}
L_{total} &= \alpha_1 * L_1(rnn_{feat}, W_1, b_1) + \\
		  &= \alpha_2 * L_2(cnn_{feat},W_2, b_2) + \\
          &= L_{fus}([cnn_{feat}, rnn_{feat}], W_3, b_3)
\end{align}

where 
\begin{align}
L_i(feat, W_i, b_i) &= L_i( Y, SoftMax(feat \cdot W_i + b_i))
\end{align}

$Y$ is the true value of the class variable. $L_1(rnn_{feat}, W_1, b_1)$ (resp. $L_2(cnn_{feat},W_2, b_2)$) is the cost function of the first (resp. the second) auxiliary classifier which takes as input the set of descriptors returned by a specific module (CNN or RNN) and the parameters to make the prediction ($W_1,b_1$ or $W_2,b_2$). $L_{fus}(cnn_{feat}, rnn_{feat}, W_3, b_3)$ is the cost function of the classifier that uses the total set of features ([$cnn_{feat},rnn_{feat}$]). This last cost function is parameterized through $W_3$ et $b_3$.
Each of the cost function is modeled through categorical cross entropy, a typical choice for multi-class supervised classification tasks~\cite{IencoGDM17}.

$L_{total}$ is optimized End-To-End.
Once the network has been trained, the prediction is carried out using only the classifier involving $W_3$ and $b_3$ which uses all the features learned by the two branches.
The cost functions $L_1$ et $L_2$, as highlighted in~\cite{HouLW17}, operate a kind of regularization that forces, within the network, the extracted features to be discriminative independently.

%% file: experiments.tex
In this section, we present the experimental setting we used and we discuss the results obtained on the data introduced in Section~\ref{sec:data}.

\subsection{Experimental Setting}
We compare the performances of the \method{} approach w.r.t the Random Forest classifier (\textit{RF}), which is commonly used for supervised classification in the field of remote sensing~\cite{LebourgeoisDVAB17}.

For the \textit{RF} model, we fixed the number of generated random trees at 200 with no depth limits imposed. For the Random Forest method, we used the python implementation supplied by the scikit-learn~\cite{scikit-learn} library. In order to compare these two methods, we supplied the same input data set both to \textit{RF} and \method{} model. Each example of the data set for this competitor has a size of 3\,669 which corresponds to 25 $\times$ 25 $\times$ 5 ($patch_{i}$) plus 34 $\times$ 16 ($ts_{i}$).

In our model we choose the value $d$ (number of hidden units in the recurrent unit $GRU$) equals to 1\,024. We empirically fixed $\alpha_1$ and $\alpha_2$ to 0.3.
During the learning phase, we used the Adam method~\cite{KingmaB14} to learn the model parameters with a learning rate equal to $2 \cdot 10^{-4}$. The training process is conducted over 400 epochs. The best model regarding the cost function's value is used in the test phase.

We implemented \method{} using the python Tensorflow library.
The learning phase takes about 15 hours while the classification on the test data takes about one minute on a workstation with an Intel (R) Xeon (R) CPU CPU E5-2667 CPU v4@3.20Ghz with 256 GB of RAM and TITAN X GPU.

In terms of data, we divided the set into two parts, one for learning and the other to test the performances of the supervised classification methods. We used 30\% of the objects for the training phase (meaning 97\,110 pixels) while the remaining 70\% are used for the test phase (meaning 225\,638 pixels). We impose that pixels of the same object belong exclusively to the train or to the test set.~\cite{IngladaVATMR17}. 
The values were normalized in the interval $[0,1]$ by spectral band.  

The assessment of classification performance is done by global precision (\textit{Accuracy}) and \textit{F-Measure}  metrics~\cite{IencoGDM17}.

\subsection{Quantitative Results}
\begin{figure}[!ht]
\centering
\includegraphics[scale=1]{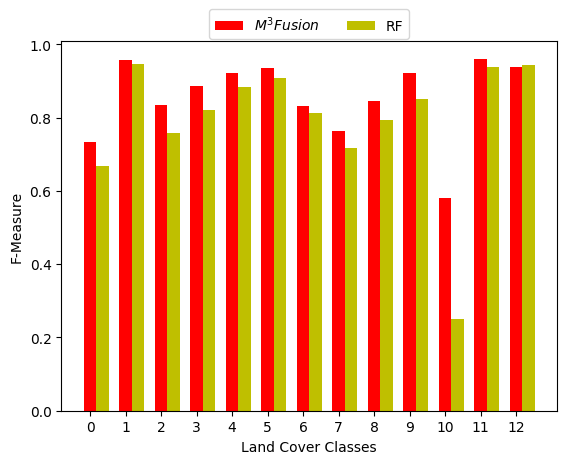}
\caption{ \label{fig:overview}  F-Measure by class of the two classification methods.}
\end{figure}

Figure~\ref{fig:overview} shows the results from both classification models in terms of F-Measure per class. We can observe that \method reaches average better performances compared to the \textit{RF}. The only exception is supplied by class (12) where performance are more than comparable and \textit{RF} obtains slightly better results.
Considering classes (1),(3),(4),(5),(7),(8) and (9), which we can consider more difficult to manage since absolute performances are low, the behavior of the deep learning method is always superior to the one exhibited by \textit{RF}. 

A class where we can notice a sensible enhancement is the class (10) \textit{Greenhouse crops}. For an \textit{F-Measure} of 0.25 given by \textit{RF}, \method{} achieves an \textit{F-Measure} of 0.58. Indeed, both the appearance and dynamics of the elements in this class are very similar to those of the built-up class. Probably, the spatial information 
extracted by the deep approach makes possible to derive more discriminative characteristics.

\begin{figure}[!ht]
\subfloat[\label{fig:cm_rf}]
{\includegraphics[width=.5\linewidth]{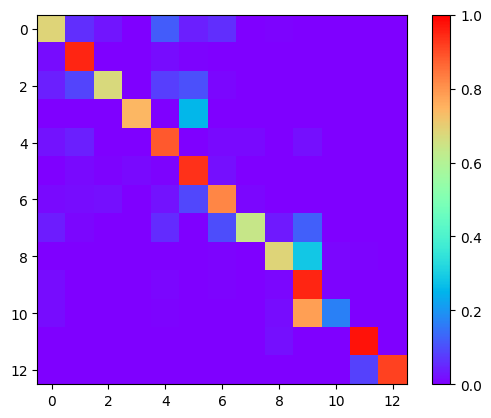} }
\subfloat[\label{fig:cm_deep}]
{\includegraphics[width=.5\linewidth]{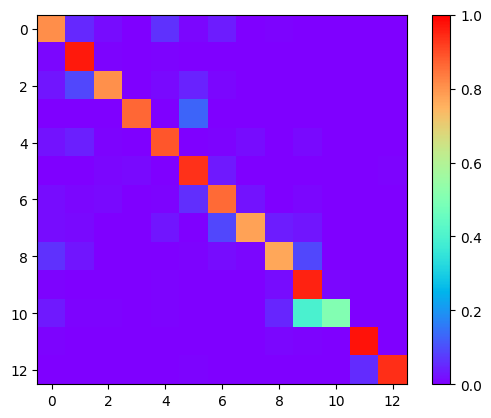}}
\caption{ Confusion matrix of \textit{Random Forest} (a) and \method{} (b).}
\end{figure}

Regarding the Accuracy results, \method{} (resp. \textit{RF}) reach a score rate of 90.67\% (resp. 87.39\%). For a more detailed analysis, we show in Figure~\ref{fig:cm_rf} (resp. Figure~\ref{fig:cm_deep}) the heat map associated with the method's confusion matrix \textit{RF} (resp. \method{}). 
We can observe that the heat map gives a good overview on the behavior of the two methods. First of all, we can observe that the \textit{Random Forest} is noisier, especially on the diagonal. This noise locates the classifier's errors in its decision. This behavior is particularly evident in class (10) where the majority of the elements of this class are categorized as class (9).
About the \method{} method, we can observe a more coherent structure along the diagonal of the heat map, which indicate less noise. Talking about class (10), also our method tends to make some confusion between classes (9) and (10), but this phenomenon is less evident in relation to the \textit{Random Forest} method since most of the elements are better classified.

The results presented so far are linked to a single split 30\%/70\% of our data set. It is known that, depending on the split of the data, the performances of the different methods may vary because simpler or more difficult examples can be involved in the train or test set. With the objective of a first understanding of the robustness of our method, regarding this phenomenon, we produced four other splits of the dataset, using the same protocol. The results of the five splits are shown in the Table~\ref{tab:decoupages}.
We can observe that the behavior of the two methods, in relation to the different splits, is similar: both methods obtain the best performance on the second split and the worst results on the third split. This behavior is related to the phenomenon we mentioned earlier.
On the other hand, we can see that \method{} always gets the best results on all splits with an Accuracy gain (resp. en F-Measure) varying between 2.28 and 4.04 (resp. 2.65 and 4.53). We can highlight two more important facts, \method{} seems to be more stable than the \textit{Random Forest} method. We observe this behavior on split number 3 where the performance of the propositional classifier decreases by more than 3 points compared to its best result. Considering \method{}, the difference between the best and worst score is around 2 points. Finally, we can note that for the results relating to the deep learning method, the difference between the Accuracy value and the F-Measure value is minimal, the two values are always rather similar and aligned. On the contrary, for the \textit{Random Forest} method we can see some discrepancy between the two measures. Accuracy measurement is always about half a point higher. Looking closely to the results, we found that \textit{RF} seems to be more influenced by the class imbalance giving more chance in its decision to the majority classes.
 
\begin{table}
\centering
\begin{tabular}{|l||c|c|||c|c|||c|c|} \hline
\textbf{Essai} & \multicolumn{2}{|c|}{ \textit{RF} } & \multicolumn{2}{|c|}{\method} & \multicolumn{2}{|c|}{\textit{Gain}} \\ \hline
& \textit{Acc.} & \textit{F-Meas.} & \textit{Acc.} & \textit{F-Meas.} 
& \textit{Acc.} & \textit{F-Meas.} \\ \hline \hline
1 & 87.39 & 87.11 & 90.67 & 90.67 & +3.28  & +3.56\\ \hline
2 & 88.47 & 88.05 & 91.52 & 91.39 &  +3.05 & +3.34 \\ \hline
3 & 85.21 & 84.62 &  89.25 & 89.15 &  +4.04 & +4.53 \\ \hline
4 & 88.33 & 88.05 & 90.61 & 90.7 &  +2.28 & +2.65 \\ \hline
5 & 87.29 & 86.88 & 90.09 & 89.96 &  +2.8 & +3.08 \\ \hline
\end{tabular}
\caption{Accuracy and F-Measure of the two methods on five different random splits of the data set \label{tab:decoupages}}
\end{table}

\subsection{Qualitative Results}
\begin{figure}[!ht]
\includegraphics[width=.99\columnwidth]{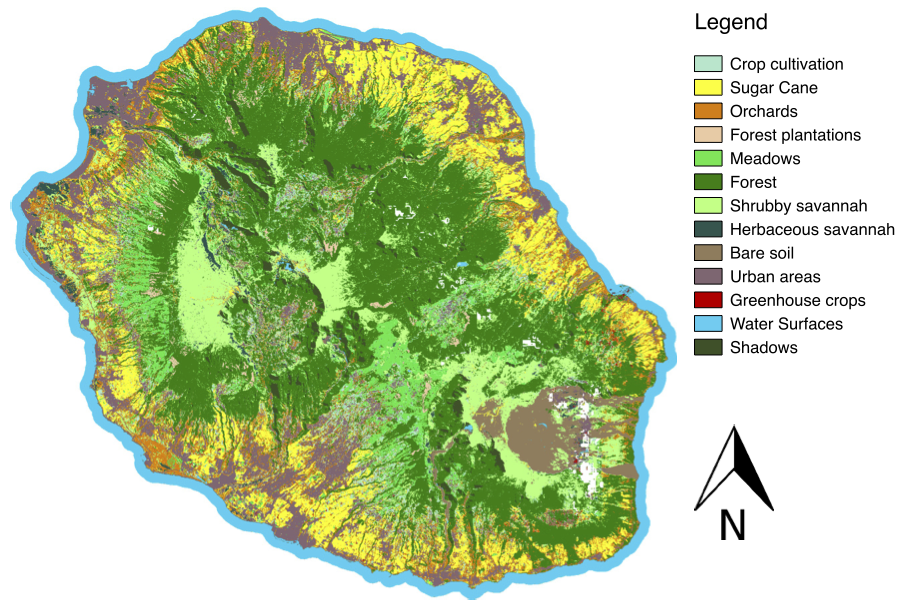}
\caption{ Map produced by \method 
}
\label{fig:map_deep}
\end{figure}

\begin{figure}[!ht]
\centering
\begin{tabular}{ccc}
\includegraphics[width=0.3\columnwidth]{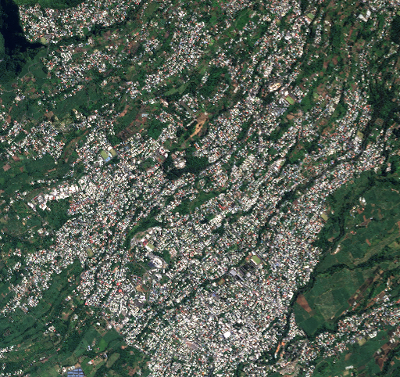} & \includegraphics[width=0.3\columnwidth]{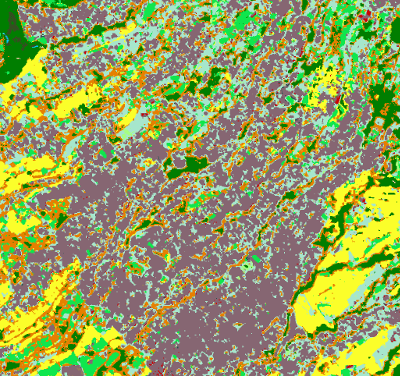} &
\includegraphics[width=0.3\columnwidth]{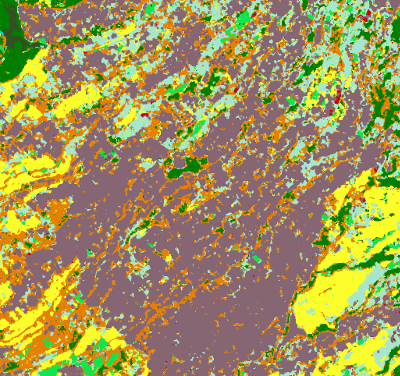} \\
\includegraphics[width=0.3\columnwidth]{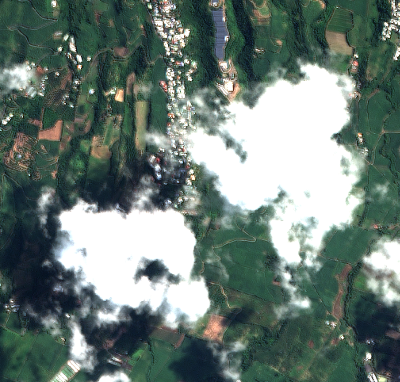} & \includegraphics[width=0.3\columnwidth]{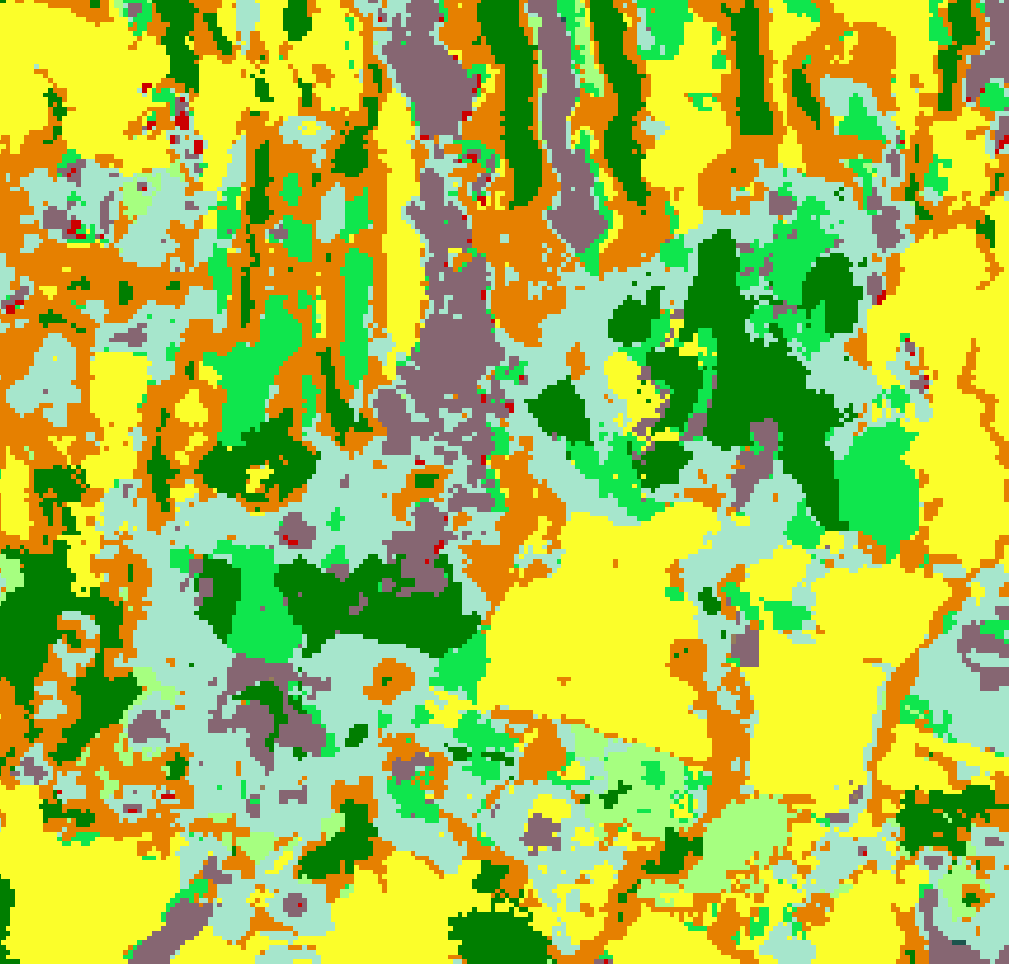} &
\includegraphics[width=0.3\columnwidth]{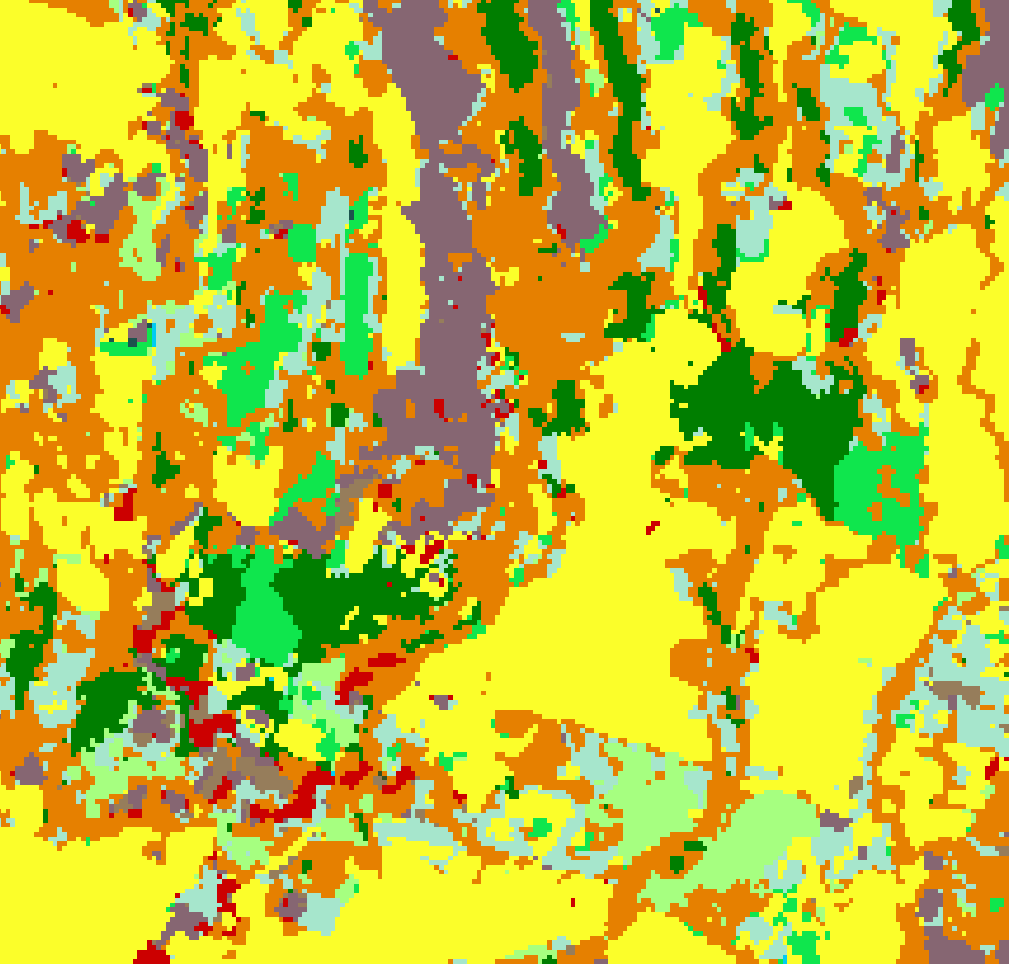}
\end{tabular}
\caption{Classification results obtained with \textit{Random Forest} and \method : right to left, excerpt from SPOT6/ image7, classification by \textit{RF}, classification by \method.}
\label{fig:details}
\end{figure}

In addition to the numerical evaluations reported in the previous section, we also propose a first qualitative evaluation of the map produced by \method. The map obtained by the \method~is also shown in Figure~\ref{fig:map_deep} for a qualitative overview. The recognition of the majority classes, i. e. the areas cultivated with sugar cane on the coast, as well as the various degraded natural areas (grasslands, savannas and forests) and the urban fabric, seem to be well localized and regular, with a less important presence of noise than on the map obtained by \textit{Random Forest} (not reported for brevity).

Some comparisons between the two maps are provided at the scale of some remarkable details in Figure~\ref{fig:details}: in the line above, a fragment of urban areas is displayed, where the presence of noise is particularly marked for the \textit{RF} map (in the middle), as shown by the transition zones between buildings that are often interpreted as market gardening, an effect that does not occur on the \method map (right). A particular interesting effect concerns the artifacts of the \textit{RF} map due to the presence of clouds or shadows (detail on the bottom line) on the VHSR image, which are not produced with the proposed method: this is probably due to an \textit{RF} bias in favor of information from the VHSR, a situation that does not occur with the proposed approach.